\documentclass{article}

\usepackage[margin=1in]{geometry}
\usepackage{amsmath,amssymb,amsfonts,bm}
\usepackage{booktabs}
\usepackage{array}
\usepackage{tabularx}
\usepackage{graphicx}
\usepackage[nolist,printonlyused]{acronym}
\usepackage[authoryear,round]{natbib}
\usepackage{hyperref}
\usepackage{multirow}
\usepackage{placeins}
\usepackage[table]{xcolor}

\newcommand{\R}{\mathbb{R}}
\newcommand{\Ad}{\mathrm{Ad}}
\newcommand{\Exp}{\mathrm{Exp}}

\newcommand{\T}{^\top}
\newcommand{\what}[1]{\widehat{#1}}

\newcommand{\nRb}{R^N_B}
\newcommand{\nPb}{p^N_B}
\newcommand{\nVb}{v^N_B}
\newcommand{\nXb}{X^N_B}

\newcommand{\SOThree}{\mathrm{SO}(3)}

\newcommand{\SEKThree}[1]{\mathrm{SE_{#1}}(3)}

\newcommand{\gyro}{\omega^B}
\newcommand{\acc}{a^B}
\newcommand{\gravity}{g^N}

\title{Four Simple Proprioceptive Estimators for Legged Robots}
\author{Frank Dellaert, Chiyun Noh, Varun Agrawal, and Ayoung Kim}
\date{}

\begin{document}

\maketitle
\begin{acronym}[ExtendedPose3]
\acro{EKF}{extended Kalman filter}
\acro{IEKF}{iterated extended Kalman filter}
\acro{InEKF}{invariant extended Kalman filter}
\acro{IMU}{inertial measurement unit}
\end{acronym}

\begin{abstract}
Legged robots carry an \acs{IMU}, but the inertial solution drifts because consumer-grade IMUs are noisy. 
However, the feet create intermittent contacts with the environment that can be used to mitigate that drift.
This report develops a sequence of increasingly expressive legged robot state estimators that leverage this.
In all cases, the floating-base state comprises attitude, position, velocity, and IMU biases.
To model foot contacts, we start from the contact-aided invariant \acs{EKF} of \citet{Hartley20ijrr_invariant}, albeit at a reduced contact update rate.
This is then augmented by replacing the measurement update by a small factor graph.
Finally, we turn the same factors into a fixed-lag smoother with contact-episode footholds, with and without an evolving \acs{IMU} bias.
To facilitate reproducibility and further research in proprioceptive legged odometry, all four variants are available in GTSAM~\citep{Dellaert22gtsam}, and we additionally provide a ROS2-compatible implementation.

  \smallskip
  \noindent\textbf{Code:}
  \href{https://github.com/borglab/gtsam/blob/develop/examples/LeggedEstimatorReplayExample.cpp}{\texttt{LeggedEstimatorReplayExample.cpp}},
  \href{https://github.com/ChiyunNoh/GTSAM-Legged-Estimator-ROS2}
  {\texttt{GTSAM-Legged-Estimator-ROS2}}.
\end{abstract}

\section{Introduction}

Proprioceptive state estimation for legged robots requires the combination of high-rate
inertial information with
intermittent geometric contact constraints.
An inexpensive, consumer-grade \acs{IMU} gives high-rate motion information, but at the
cost of integration drift.
A foot in rigid contact with the environment gives a strong geometric constraint, but
only while that foot is actually in stance.
A useful proprioceptive estimator combines both information sources to limit drift, creating a strong foundation to optionally fuse with exteroceptive sensor information.

Our estimators are deliberately simpler than several related legged-estimation systems in the literature.
In contrast to the estimators of \citet{Bloesch13rss_legged}, \citet{Hartley20ijrr_invariant}, and our own kinematic-chain factor-graph
estimator~\citep{Agrawal22humanoids_kinematic}, we do not use joint angles as measurements.
Instead, we assume that forward kinematics has been computed by a front end, and the estimator receives a 3D foot point in the body frame.
This also means that the contact-measurement uncertainty is not informed by the
forward-kinematics Jacobians.
Finally, we do not run contact updates at the joint-angle measurement rate; we wait for
touchdown events or until a significant amount of time has passed.
In our experiments this event-driven update schedule appears beneficial, with noticeably less drift than a high-rate contact-update schedule.

This report follows an incremental approach, gradually increasing estimator complexity.
We start with the contact-aided invariant \acs{EKF} viewpoint of
\citet{Hartley20ijrr_invariant}, itself based on the invariant filtering paradigm of
\citet{Barrau17tac_IEKF}, since it provides a clean representation of contact-aided inertial navigation on a Lie group.
In a second filter, we keep the same prediction step but replace the sequential
measurement update by a small factor graph so simultaneous contacts are solved together
and possibly iterated upon.
Third, once the update has become a graph, we let the graph span a short time window;
this results in a fixed-lag smoother, and it forces footholds to become explicit
navigation-frame variables rather than blocks inside the current filter state.
Finally, we replace the single persistent bias in the smoother with an evolving bias
trajectory modeled as a Markov chain.

In terms of notation, we use the following conventions:
the body frame is denoted by $B$, the navigation frame by $N$, and the floating-base state
$\nXb$ is given by
\begin{equation}
  \nXb = (\nRb,\nPb,\nVb).
\end{equation}
Above $\nRb$ maps body-frame vectors to the
navigation frame, $\nPb$ is the body position in $N$, and
$\nVb$ is the body velocity in $N$. A contact measurement
$c^B_i$ is the vector from the body frame to foot $i$, expressed in the body frame.

\section{\texorpdfstring{Invariant \acs{EKF}}{Invariant EKF}}
\label{sec:invariant-ekf}
\subsection{State and local coordinates}

The invariant filter keeps attitude, translation, velocity, and footholds in a
Lie-group state with local tangent-space errors~\citep{Barrau17tac_IEKF}. This matters
because a small attitude error changes how position, velocity, and foothold
errors are interpreted. The filter mean lives on a Lie group, while the
covariance lives in a local tangent vector. That is the same practical
compromise used by ordinary error-state filters, but with error coordinates
chosen to respect the group structure.

The Lie group $\SEKThree{K}$ combines one attitude $R$ and
$K$ vector blocks. For any $K$, the semidirect-product Lie group is defined as
\begin{equation}
  \SEKThree{K}
  =
  \left\{
    \begin{bmatrix}
      R & X \\
      0_{K\times 3} & I_K
    \end{bmatrix}
    \ \middle|\
    R \in \SOThree,\ X=[x_1\ \cdots\ x_K]\in \R^{3\times K}
  \right\}.
\end{equation}
Equivalently, an element is $(R,x_1,\dots,x_K)$, with group product
\begin{equation}
  (R,x_1,\dots,x_K)(S,y_1,\dots,y_K)
  =
  (RS,\ x_1+Ry_1,\dots,x_K+Ry_K).
\end{equation}
The group only provides structure; it does not assign semantics to the vector
blocks. For $k$ feet, the invariant filter from \citep{Hartley20ijrr_invariant} uses
$\SEKThree{k+2}$, with
\begin{equation}
  X =
  (\nRb,\nPb,\nVb,f^N_1,\dots,f^N_k).
\end{equation}
Note that unlike \citet{Hartley20ijrr_invariant} we use a \textit{left-invariant} filter
convention: each foothold $f^N_i$ is stored as a point in the navigation frame, but the
corresponding perturbation is a body-frame foot error. We also simplify the math by using
the recent linear observer structure from \cite{Barrau19scl_LeftLinear}, as explained below.

\subsection{IMU-driven Prediction}

A single IMU-sample prediction decomposes inertial navigation into gravity,
autonomous drift, and body-frame \acs{IMU} motion. For one zero-order-hold
\acs{IMU} sample over a time step $\Delta t$, it has the left-linear Lie-group
form~\citep{Barrau19scl_LeftLinear,Chirikjian11book_Stochastic}
\begin{equation}
  X^+ = W\,\phi(X)\,U,
\end{equation}
with known left factor $W$, deterministic autonomous flow $\phi$, and known
right factor $U$. The notation is compact, but the pieces have simple
physical roles.

The left factor $W$ is the gravity increment in the navigation frame. For the
base $\SEKThree{2}$ state, with gravity $\gravity$ and time step $\Delta t$, it is
\begin{equation}
  W =
  \begin{bmatrix}
    I_3 & \frac{1}{2}\gravity\Delta t^2 & \gravity\Delta t \\
    0_{1\times 3} & 1 & 0 \\
    0_{1\times 3} & 0 & 1
  \end{bmatrix}.
\end{equation}
For the foot-augmented $\SEKThree{k+2}$ state, the same nonzero column vectors appear for the position and velocity slots, while the foothold columns in
$W$ are zero.

The automorphism $\phi$ is the autonomous drift that advances position via the
velocity. For the base $\SEKThree{2}$ state, we get
\begin{equation}
  \phi(\nRb,\nPb,\nVb)
  =
  (\nRb,\ \nPb+\nVb\Delta t,\ \nVb),
\end{equation}
and for the foot-augmented state it leaves the foothold columns unchanged.

The right factor $U$ is the single-sample zero-order-hold inertial increment
in the body frame. It is built from the angular velocity $\tilde\gyro$ measured from the gyroscope and the specific-force sample $\tilde \acc$ obtained from the accelerometer,
which after subtracting the current bias estimate gives:
\begin{equation}
  \gyro = \tilde\gyro - b_g,\qquad
  \acc = \tilde \acc-b_a,\qquad
  \Omega = \Exp(\gyro \Delta t)
\end{equation}
The base $\SEKThree{2}$ increment is
\begin{equation}
  U =
  \begin{bmatrix}
    \Omega & \Gamma_l(\gyro \Delta t)\,\acc\Delta t^2
    & J_l(\gyro\Delta t)\,\acc\Delta t \\
    0_{1\times 3} & 1 & 0 \\
    0_{1\times 3} & 0 & 1
  \end{bmatrix},
\end{equation}
where $J_l$ is the left $\SOThree$ Jacobian and $\Gamma_l$ is the left Gamma
matrix; their closed forms are given in Appendix~\ref{app:so3-matrices}.
For the foot-augmented state, the foothold columns in $U$ are zero.
The filter prediction uses one-sample integration rather than preintegration: we perform this update at the \acs{IMU} rate.
The smoothers introduced later will instead accumulate many such samples between update steps.

The left-linear form gives a compact state-independent prediction Jacobian. Let
\begin{equation}
  \hat X^+ = W\,\phi(\hat X)\,U.
\end{equation}
In the left-invariant formulation, we have
\begin{equation}
  \phi(\hat X\Exp(\xi)) \approx \phi(\hat X)\Exp(\Phi\xi),
\end{equation}
where $\Phi$ is the differential of the autonomous flow $\phi$ at
the identity: it is identity except for the
position-velocity block $\Phi_{pv}=\Delta t\,I_3$.
The Jacobian of composition is given by an Adjoint, hence
\begin{equation}
  X^+ \approx \hat X^+\Exp(\Ad_{U^{-1}}\Phi\xi),
\end{equation}
which gives the prediction Jacobian as
\begin{equation}
  \boxed{
    \xi^+ \approx A\,\xi,
    \qquad
    A := \Ad_{U^{-1}}\Phi.
  }
\end{equation}

We model the footholds as stationary during prediction, but the corresponding
foot-error block is not the identity $I_3$ because the left-invariant error is a
right perturbation on the semidirect product.
Writing $(\hat \nRb)^+ = \hat \nRb\Delta R$, the foothold block is
\begin{equation}
  \boxed{
    A_{f_i f_i} =
    \frac{\partial(\delta f^B_i)^+}{\partial \delta f^B_i}
    = \Delta R\T,
  }
\end{equation}
This indicates the body-frame coordinate chart used by the covariance rotates with the nominal body.
The derivation is given in Appendix~\ref{app:foothold-transport}.

\subsection{Contact Update}

Following the approach proposed by \citet{Bloesch13rss_legged}, contact turns
a stance foot into a temporary landmark observed from the moving
body. If foot $f_i$ is in stance, the measured vector $z^B_i$ gives the foot position relative to the body position $\nPb$, expressed in the body frame.
During one stance phase the foot is approximately fixed in the navigation frame, and the robot
repeatedly observes it from the moving body.

Using the left-invariant formulation, the physical foot point is, to first order
\begin{equation}
  f^N_i = \hat f^N_i + \hat \nRb \delta f^B_i.
\end{equation}
The contact measurement model predicts the body-frame vector from base to
foothold, i.e.,
\begin{equation}
  z^B_i = h_i(\nRb,\nPb,f^N_i)
  = (\nRb)\T(f^N_i-\nPb) + n_i.
\end{equation}
We define the contact forward kinematics measurement as
\begin{equation}
   z^B_i = (\hat \nRb)\T(\hat f^N_i-\hat \nPb).
\end{equation}
and using the body-frame foot-error chart above, the first-order contact linearization gives
\begin{equation}
  \delta z^B_i \approx
  \what{\hat z^B_i}\,\delta\phi - \delta \nPb + \delta f^B_i.
\end{equation}
The foot block in the resulting measurement Jacobian is therefore constant:
\begin{equation}
  \boxed{H_f = I_3.}
\end{equation}

\subsection{Touchdown and Liftoff}

The invariant filter has a fixed foothold slot for each leg, but a slot denotes
only the currently tracked contact episode for that leg. The contact front end
provides packets containing the stance feet, together
with a touchdown flag for any foot that has just transitioned from swing to
stance. The filter does \textit{not} apply a contact correction
for every contact packet: instead, we update only when
a packet contains at least one touchdown. When packets contain no touchdown, an
additional periodic contact update is scheduled only after at least
$\Delta t_{\max}=100\,\mathrm{ms}$ has elapsed since the previous scheduled
contact update. Between scheduled contact
updates, the filter only performs \acs{IMU} prediction.

At a scheduled contact update, any foot that was in contact at the previous scheduled update but
is absent from the current stance set is considered to have lifted off.
For that foot, the old contact episode is removed from the fixed-size state in place.
Let $l$ denote the three coordinates of the leaving foothold slot and let $r$
denote all remaining coordinates. Marginalizing out the leaving foothold gives
the retained covariance
\begin{equation}
  P^+_{rr}
  =
  \left(
    \Lambda^-_{rr}
    - \Lambda^-_{rl}(\Lambda^-_{ll})^{-1}\Lambda^-_{lr}
  \right)^{-1},
  \qquad \Lambda^- = (P^-)^{-1}.
\end{equation}
This is the marginal submatrix on the retained coordinates: we write the result
back into the same fixed foot-slot layout, zero the cross-covariances to the
vacated slot, and assign an independent foothold-initialization covariance $\sigma_f^2
I_3$ for the new
touchdown initialization:
\begin{equation}
  P^+
  =
  \begin{bmatrix}
    P^+_{rr} & 0 \\
    0 & \sigma_f^2 I_3
  \end{bmatrix},
\end{equation}
where $\sigma_f$ is very loose.
The slot mean is immediately initialized from the new touchdown measurement,
\begin{equation}
  \hat f^N_i = \hat \nPb + \hat \nRb z^B_i.
\end{equation}
After this initialization, the contact correction is applied to all active feet
in the scheduled packet. A stance foot that is not marked as touchdown is not
reinitialized; it is simply included in the contact update. Thus, after a
touchdown, the same foothold is updated again only at the next scheduled contact
update, either because some other foot has touched down or the periodic interval $\Delta t_{\max}$ has elapsed.

\subsection{Optional Height Prior}

If an external terrain-height measurement $h$ is available, it can be added as a
scalar prior on the navigation-frame foothold height. The terrain height measurement in the navigation frame can be written as
\begin{equation}
  0 = f^N_{i,z} - h + n_h.
\end{equation}
The height-prior Jacobian must account for the body-frame foot perturbation. On
the foot block it is
\begin{equation}
  H_h = e_3\T \hat \nRb.
\end{equation}
We emphasize that this height prior is optional. In the reported experiments, no terrain height is
supplied and the height-prior factor is disabled; vertical information comes
only through the inertial model and the measured contact kinematics.

\section{Replacing the Update by a Local Graph}
\label{sec:graph-update-filter}

\begin{figure}[htbp]
  \centering
  \includegraphics[width=0.72\textwidth]{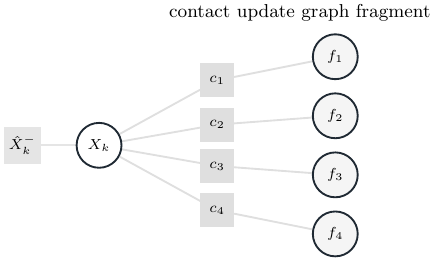}
  \caption{Local graph update for the invariant filter. The predicted
    semidirect-product state $X_k$ is constrained by a prior factor
    $\hat X_k^{-}$, while the active contact factors $c_i$ connect the same
  state to the currently tracked footholds $f_i$.}
  \label{fig:local-graph-update}
\end{figure}

Our next filter replaces the sequential \acs{EKF} correction with one small
batch correction at the current time. The prediction above is already in a
convenient filter form, so only the measurement phase changes. At a single time
step, the predicted current-state belief can be converted into a small
nonlinear graph. The prior factor represents the filter belief after
prediction; the contact and (optional) height factors represent the measurements arriving
at the same time. In other words, the estimator is still a filter from the
outside, but the correction step is now a tiny batch optimization:
\begin{itemize}
  \item One prior factor on the current semidirect-product state, using the
    filter mean and covariance;
  \item One contact factor for each active foot;
  \item Optional height-prior factors.
\end{itemize}
We run one local Levenberg--Marquardt optimization solve for the correction step,
initialized at the predicted state, and then write the optimized mean and
marginal covariance back into the filter.

The local graph update preserves the invariant filter state and prediction
model. The state, prediction model, contact model, and body-frame foot-error
chart are the same as in the sequential invariant filter. The benefit is that
simultaneous contacts are handled symmetrically in one local nonlinear solve,
rather than through an arbitrary sequence of \acs{EKF} updates.
Figure~\ref{fig:local-graph-update} shows the resulting graph fragment.

\section{From Local Updates to a Fixed-lag Smoother}
\label{sec:fixed-lag-single}

The fixed-lag smoother extends the local graph idea in~\ref{sec:graph-update-filter} from one time step to a
short recent history. Once the contact update is expressed as factors, it is
natural to keep more than the current state. While a
filter summarizes the past by one Gaussian posterior density at the current time, a
smoother keeps a short recent history alive so delayed or repeated contact
information can influence several nearby base states before marginalization.

The smoother needs contact-episode footholds rather than fixed filter foot
slots. The semidirect-product filter stores all currently labeled feet in one
current-state variable, which is appropriate for a fixed-size filter. A
smoother, however, needs to represent contact \textit{episodes}: if a foot lifts off and
later touches down somewhere else, those are different physical footholds. The
smoother therefore uses explicit navigation-frame landmark variables for
footholds:
\begin{equation}
  f^N_{i,e},
\end{equation}
where $i$ is the foot label and $e$ is the contact episode.

\subsection{Event-based Timeline}

An event-based timeline keeps the smoother small by creating base states only
at contact events, defined as before by footfalls or by a fixed elapsed time.
Unlike the filter, the smoother uses the preintegrated \acs{IMU} factor from
\cite{Forster17tro_preintegration} to accumulate the high-rate \acs{IMU} samples between contact
events. If the interval from event
$i$ to event $j$ contains many \acs{IMU} samples, preintegration summarizes
them as a single relative $\SEKThree{2}$ increment
\begin{equation}
  \Delta U_{ij} =
  \begin{bmatrix}
    \Delta R_{ij} & \Delta p_{ij} & \Delta v_{ij} \\
    0_{1\times 3} & 1 & 0 \\
    0_{1\times 3} & 0 & 1
  \end{bmatrix}.
\end{equation}
The preintegrated increment is the smoother analogue of the filter's
one-sample $U$. Here $\Delta R_{ij}$, $\Delta p_{ij}$, and
$\Delta v_{ij}$ are accumulated from the sequence of body-frame gyroscope and
specific-force samples over the interval, around a chosen bias linearization
point. With a single persistent bias, the window contains:
\begin{itemize}
  \item One $(\nRb,\nPb,\nVb)$ base state per contact event;
  \item One shared \acs{IMU} bias variable;
  \item One navigation-frame foothold landmark per active contact episode.
\end{itemize}
The smoother can still provide a current estimate between contact events by dead-reckoning
forward from the latest optimized base state.

\subsection{Factors}
\begin{figure}[htbp]
  \centering
  \includegraphics[width=0.82\textwidth]{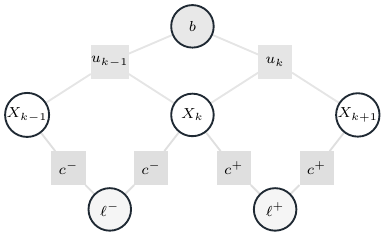}
  \caption{Fixed-lag smoother with a single persistent bias variable. Each
    $X$ node is a contact-event base state, $b$ is shared across the window,
    $u$ factors are preintegrated inertial constraints, and $c$ factors attach
  event states to contact-episode foothold landmarks $\ell$.}
  \label{fig:fixed-lag-single-bias}
\end{figure}

Figure~\ref{fig:fixed-lag-single-bias}
shows the graph structure for this single-bias smoother.
The single-bias fixed-lag graph combines inertial, bias, contact, and height
factors:
\begin{itemize}
  \item A preintegrated inertial factor between successive contact-event base
    states;
  \item An initial prior on the persistent bias variable;
  \item Contact factors between event base states and active foothold episode
    landmarks;
  \item Optional height priors on foothold landmarks.
\end{itemize}
The smoother contact factor keeps the same residual but changes the foothold
variable representation. The contact residual is still
\begin{equation}
  z^B_i = (\nRb)\T(f^N_{i,e}-\nPb) + n_i,
\end{equation}
but now $f^N_{i,e}$ is an explicit navigation-frame point variable, not a
body-frame foot-error block inside the current filter state. A new touchdown
creates a new landmark key. Old contact episodes are properly marginalized when they
fall outside the fixed-lag window.

Note that in the filter the dynamics were invariant, which allowed for proper covariance
propagation even when the state estimate was imperfect. In the smoother, while still using
a left-invariant error, the footholds are no longer part of a semidirect-product state
with the floating base. Instead, we rely on optimization to compute the posterior density
over the window after convergence.

\section{Letting the Bias Evolve}
\label{sec:fixed-lag-combined}
\begin{figure}[htbp]
  \centering
  \includegraphics[width=0.78\textwidth]{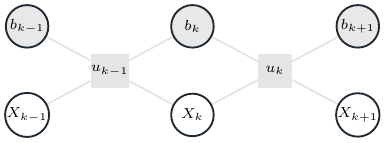}
  \caption{Evolving-bias inertial structure. The combined preintegrated factors
    connect neighboring base states and neighboring bias states, replacing the
  single shared-bias assumption in Figure~\ref{fig:fixed-lag-single-bias}.}
  \label{fig:fixed-lag-combined-bias}
\end{figure}

Finally, the evolving-bias smoother relaxes the single-bias assumption while keeping the
same contact-episode structure. The previous smoother keeps one persistent bias
variable, which is compact and often useful for short experiments, but
it assumes the bias is constant across the entire run. That assumption can be too strong for
longer experiments, changing thermal conditions, or low-cost inertial sensors. The
combined fixed-lag smoother therefore represents each contact event with
separate pose, velocity, and bias variables:
\begin{equation}
  (T^N_{B,j}, v^N_{B,j}, b_j).
\end{equation}
Successive events are connected by an extended preintegrated inertial factor,
which also models bias evolution as a random walk.

\section{Experimental Evaluation}
\subsection{Dataset and Metrics}

\begin{figure}[htbp]
  \centering
  \includegraphics[width=0.52\textwidth]{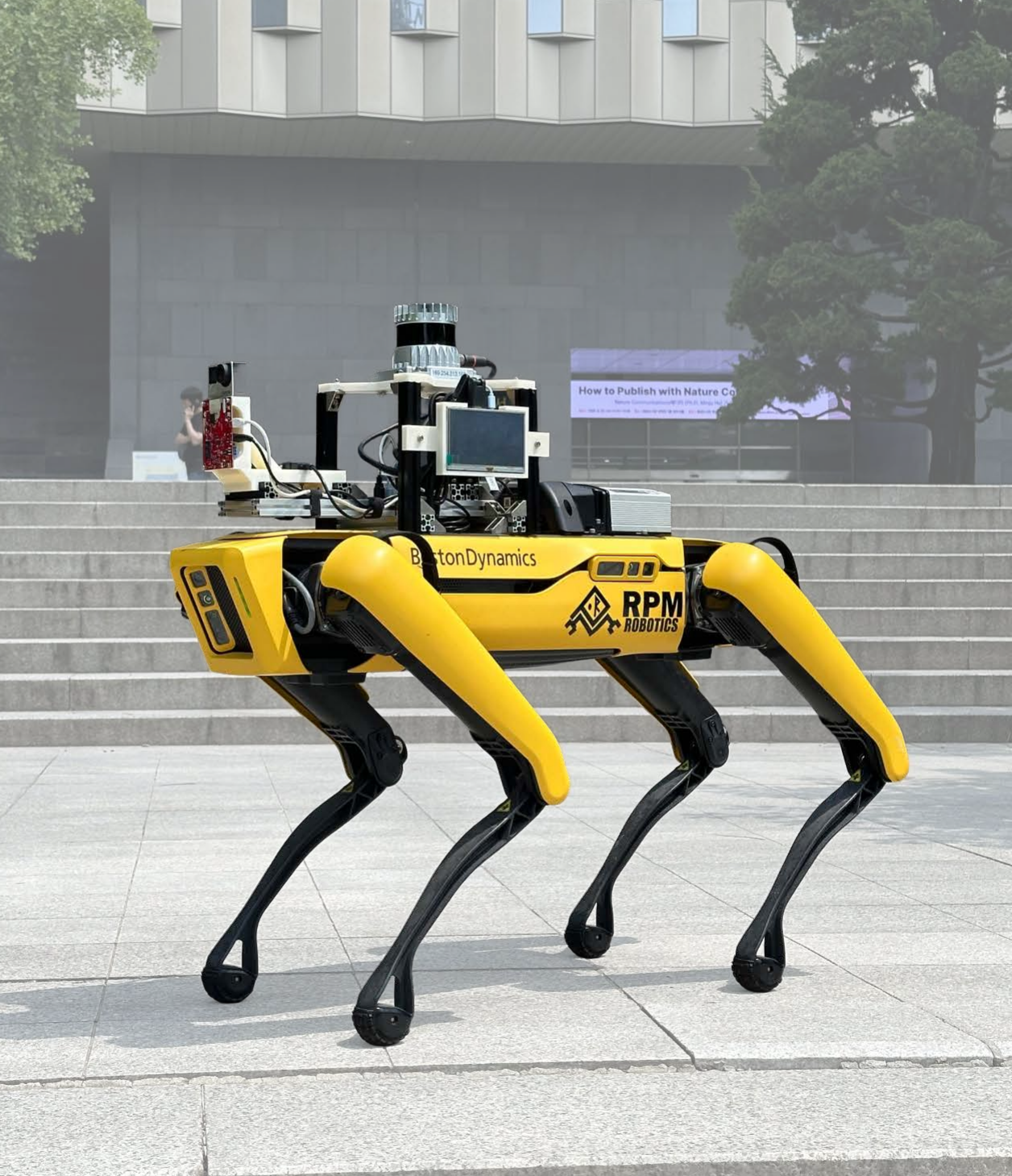}
  \caption{Experimental platform used in the GaRLILEO dataset.}
  \label{fig:platform}
\end{figure}

We evaluate the estimators on the GaRLILEO dataset~\citep{Chiyun25_garlileo}, which was
collected using a Boston Dynamics Spot platform with leg joint encoders, foot contact
measurements, an \acs{IMU}, radar, and reference sensors for ground-truth generation, as
shown in Fig.~\ref{fig:platform}. Since this note focuses on proprioceptive legged state
estimation, the evaluated GTSAM
variants use only inertial and leg measurements, without radar or LiDAR input. All
experiments employed the calibrated IMU-to-body transform provided with the GaRLILEO
dataset\footnote{All four estimator implementations support this additional transform}.

We use the SNU sequences of the dataset, which cover indoor and outdoor environments
including corridors, loops, stairs, slopes, tunnels, and open campus spaces. Ground-truth
trajectories are provided by the dataset using a dense Terrestrial Laser Scanning (TLS)
prior map and LiDAR-inertial trajectory estimation. We divide the evaluation into
non-elevation-change sequences (Atrium, CorriLoop, BridgeLoop, and Tunnel) and
elevation-change sequences (Downstair, BiCorridor, Upstair, Quad, Overpass, and SlopeStair).

Following the evaluation protocol of GaRLILEO, we report the root mean square error
(RMSE) of absolute pose error (APE) and relative pose error (RPE), each decomposed into
translational and rotational components. To explicitly measure vertical drift, we also
report the z-axis absolute pose error, denoted as APE$_z$. Lower values are better for
all metrics. For each metric in each sequence, the best result is highlighted in bold
green, and the second-best result is underlined. The four GTSAM variants evaluated in the
following sections correspond to the estimators introduced in
Sections~\ref{sec:invariant-ekf}--\ref{sec:fixed-lag-combined}. \textbf{Inv. EKF} denotes
the invariant filter, \textbf{Inv. IEKF} denotes the graph-update filter,
\textbf{FL-Single} denotes the fixed-lag smoother with a single persistent bias, and
\textbf{FL-Combined} denotes the combined fixed-lag smoother with an evolving bias.

\subsection{Comparison with Existing Legged-IMU Odometry}

\providecommand{\best}[1]{\cellcolor{green!25}\textbf{#1}}
\providecommand{\second}[1]{\underline{#1}}

\begin{table}[htbp]
\centering
\scriptsize
\setlength{\tabcolsep}{5.6pt}
\renewcommand{\arraystretch}{1.08}
\caption{Quantitative comparison on non-elevation-change (2D) sequences.}
\label{tab:legged_odometry_non_elevation}
{%
\begin{tabular}{lllccccccc}
\toprule
\multirow{2}{*}{Sequence} & \multirow{2}{*}{Metric} & \multirow{2}{*}{Unit} & \multicolumn{3}{c}{Other legged odometry} & \multicolumn{4}{c}{GTSAM legged estimators} \\
\cmidrule(lr){4-6}\cmidrule(lr){7-10}
 & & & Pronto & MUSE & Holistic & Inv. EKF & Inv. IEKF & FL-Single & FL-Combined \\
\midrule
\multirow{5}{*}{CorriLoop} & \multirow{2}{*}{APE RMSE} & Trans. (m) & 8.988 & 5.811 & 14.780 & \second{2.003} & \best{1.997} & 2.010 & 2.296 \\
 &  & Rot. (deg) & 25.691 & 18.871 & 25.942 & 1.802 & \second{1.761} & 1.815 & \best{0.994} \\
 & \multirow{2}{*}{RPE RMSE} & Trans. (m) & 0.200 & 0.127 & 0.173 & 0.097 & 0.097 & \best{0.096} & \second{0.096} \\
 &  & Rot. (deg) & \best{0.711} & 1.028 & 1.358 & 0.781 & 0.781 & \second{0.740} & 0.808 \\
 & APE Vertical & Trans. (m) & 1.519 & 2.514 & 12.668 & 0.533 & \second{0.529} & \best{0.525} & 1.776 \\
\midrule
\multirow{5}{*}{Atrium} & \multirow{2}{*}{APE RMSE} & Trans. (m) & 7.849 & 3.297 & 7.407 & 1.527 & 1.528 & \second{1.515} & \best{0.993} \\
 &  & Rot. (deg) & 21.448 & 13.597 & 17.631 & 1.935 & 1.926 & \second{1.790} & \best{1.092} \\
 & \multirow{2}{*}{RPE RMSE} & Trans. (m) & 0.293 & \best{0.090} & 0.151 & 0.106 & 0.106 & 0.105 & \second{0.090} \\
 &  & Rot. (deg) & 0.700 & 0.925 & 1.209 & \second{0.556} & \best{0.555} & 0.560 & 0.565 \\
 & APE Vertical & Trans. (m) & 0.592 & 0.685 & 5.303 & 0.480 & 0.484 & \second{0.407} & \best{0.328} \\
\midrule
\multirow{5}{*}{BridgeLoop} & \multirow{2}{*}{APE RMSE} & Trans. (m) & 6.684 & 4.195 & 5.037 & 1.989 & \second{1.988} & 2.014 & \best{1.695} \\
 &  & Rot. (deg) & 16.739 & 26.513 & 25.208 & 4.289 & \second{4.123} & 4.367 & \best{2.218} \\
 & \multirow{2}{*}{RPE RMSE} & Trans. (m) & 0.256 & 0.200 & 0.162 & 0.134 & \second{0.133} & 0.135 & \best{0.110} \\
 &  & Rot. (deg) & 0.950 & 0.967 & 1.823 & 0.806 & \second{0.803} & 0.830 & \best{0.782} \\
 & APE Vertical & Trans. (m) & 5.160 & 2.250 & 3.870 & \best{0.793} & 0.809 & \second{0.802} & 1.163 \\
\midrule
\multirow{5}{*}{Tunnel} & \multirow{2}{*}{APE RMSE} & Trans. (m) & 6.171 & 10.194 & 28.576 & 5.443 & 5.483 & \second{5.412} & \best{4.255} \\
 &  & Rot. (deg) & 13.369 & 11.199 & 41.034 & 3.464 & 3.460 & \second{3.223} & \best{1.423} \\
 & \multirow{2}{*}{RPE RMSE} & Trans. (m) & 0.158 & 0.105 & 0.199 & \second{0.086} & 0.087 & 0.088 & \best{0.079} \\
 &  & Rot. (deg) & 0.482 & 0.500 & 0.903 & 0.441 & \second{0.436} & 0.439 & \best{0.425} \\
 & APE Vertical & Trans. (m) & 4.161 & 5.425 & 9.741 & \second{1.991} & 2.018 & 2.052 & \best{0.351} \\
\bottomrule
\end{tabular}%
}
\vspace{1mm}
\end{table}

\providecommand{\best}[1]{\cellcolor{green!25}\textbf{#1}}
\providecommand{\second}[1]{\underline{#1}}

\begin{table}[htbp]
\centering
\scriptsize
\setlength{\tabcolsep}{5.0pt}
\renewcommand{\arraystretch}{1.08}
\caption{Quantitative comparison on elevation-change (3D) sequences.}
\label{tab:legged_odometry_elevation}
{%
\begin{tabular}{lllccccccc}
\toprule
\multirow{2}{*}{Sequence} & \multirow{2}{*}{Metric} & \multirow{2}{*}{Unit} & \multicolumn{3}{c}{Other legged odometry} & \multicolumn{4}{c}{GTSAM legged estimators} \\
\cmidrule(lr){4-6}\cmidrule(lr){7-10}
 & & & Pronto & MUSE & Holistic & Inv. EKF & Inv. IEKF & FL-Single & FL-Combined \\
\midrule
\multirow{5}{*}{Downstair} & \multirow{2}{*}{APE RMSE} & Trans. (m) & 11.613 & 8.323 & 30.346 & 4.575 & 4.591 & \second{4.544} & \best{3.842} \\
 &  & Rot. (deg) & 10.791 & 8.595 & 43.444 & \second{1.585} & 1.592 & \best{1.536} & 1.829 \\
 & \multirow{2}{*}{RPE RMSE} & Trans. (m) & 0.187 & 0.182 & 0.174 & 0.133 & 0.133 & \second{0.132} & \best{0.116} \\
 &  & Rot. (deg) & \second{0.658} & 1.095 & 1.637 & 0.713 & 0.711 & 0.679 & \best{0.614} \\
 & APE Vertical & Trans. (m) & 2.490 & 4.843 & 5.601 & \second{1.124} & 1.130 & \best{1.025} & 1.317 \\
\midrule
\multirow{5}{*}{BiCorridor} & \multirow{2}{*}{APE RMSE} & Trans. (m) & 15.078 & 15.035 & 13.990 & \second{2.046} & \best{2.034} & 2.082 & 2.602 \\
 &  & Rot. (deg) & 44.116 & 51.048 & 38.014 & 1.607 & \best{1.563} & \second{1.605} & 2.307 \\
 & \multirow{2}{*}{RPE RMSE} & Trans. (m) & 0.296 & 0.152 & 0.216 & \best{0.099} & 0.100 & 0.100 & \second{0.099} \\
 &  & Rot. (deg) & 0.840 & 0.918 & 1.419 & 0.680 & \best{0.678} & 0.684 & \second{0.678} \\
 & APE Vertical & Trans. (m) & 3.236 & 2.880 & 7.868 & \second{0.261} & 0.263 & \best{0.253} & 2.058 \\
\midrule
\multirow{5}{*}{Upstair} & \multirow{2}{*}{APE RMSE} & Trans. (m) & 6.238 & 5.307 & 7.238 & \second{1.460} & \best{1.456} & 1.496 & 2.118 \\
 &  & Rot. (deg) & 21.379 & 27.242 & 30.934 & 3.219 & \second{3.106} & 3.126 & \best{1.496} \\
 & \multirow{2}{*}{RPE RMSE} & Trans. (m) & 0.139 & 0.179 & 0.141 & 0.117 & 0.117 & \second{0.116} & \best{0.108} \\
 &  & Rot. (deg) & \best{0.647} & 0.979 & 1.643 & 0.750 & 0.748 & 0.728 & \second{0.681} \\
 & APE Vertical & Trans. (m) & 4.485 & 3.206 & 5.574 & \best{0.343} & \second{0.350} & 0.357 & 1.898 \\
\midrule
\multirow{5}{*}{Quad} & \multirow{2}{*}{APE RMSE} & Trans. (m) & 11.851 & 172.151 & 122.381 & 6.744 & \second{6.738} & \best{6.670} & 8.132 \\
 &  & Rot. (deg) & 6.218 & 105.269 & 79.279 & 2.665 & \second{2.660} & \best{2.385} & 3.944 \\
 & \multirow{2}{*}{RPE RMSE} & Trans. (m) & 0.157 & 0.119 & 0.152 & \second{0.087} & 0.087 & 0.088 & \best{0.077} \\
 &  & Rot. (deg) & 0.543 & 1.709 & 0.969 & 0.526 & 0.526 & \best{0.512} & \second{0.522} \\
 & APE Vertical & Trans. (m) & 2.924 & 6.519 & 4.012 & 2.840 & 2.846 & \second{2.581} & \best{2.207} \\
\midrule
\multirow{5}{*}{Overpass} & \multirow{2}{*}{APE RMSE} & Trans. (m) & \second{4.039} & 6.507 & 5.920 & 7.627 & 7.664 & 7.755 & \best{2.604} \\
 &  & Rot. (deg) & 5.367 & 10.373 & 16.034 & 4.331 & 4.311 & \best{4.124} & \second{4.836} \\
 & \multirow{2}{*}{RPE RMSE} & Trans. (m) & 0.139 & 0.191 & \second{0.125} & 0.207 & 0.207 & 0.207 & \best{0.110} \\
 &  & Rot. (deg) & \best{0.582} & 0.972 & 1.594 & 0.819 & 0.815 & 0.772 & \second{0.771} \\
 & APE Vertical & Trans. (m) & \second{1.227} & 2.442 & 2.424 & 7.066 & 7.105 & 7.150 & \best{0.676} \\
\midrule
\multirow{5}{*}{SlopeStair} & \multirow{2}{*}{APE RMSE} & Trans. (m) & 57.862 & 35.519 & 30.420 & 3.292 & 3.282 & \best{3.248} & \second{3.523} \\
 &  & Rot. (deg) & 67.826 & 54.144 & 40.790 & 1.177 & \second{1.161} & 1.164 & \best{1.127} \\
 & \multirow{2}{*}{RPE RMSE} & Trans. (m) & 0.700 & 0.151 & 0.205 & 0.097 & 0.097 & \second{0.098} & \best{0.093} \\
 &  & Rot. (deg) & 0.954 & 1.063 & 1.274 & \second{0.694} & \best{0.693} & 0.721 & 0.748 \\
 & APE Vertical & Trans. (m) & 3.607 & 0.967 & 8.427 & 0.590 & \second{0.590} & \best{0.519} & 2.008 \\
\bottomrule
\end{tabular}%
}
\vspace{1mm}
\end{table}

Table~\ref{tab:legged_odometry_non_elevation} and
Table~\ref{tab:legged_odometry_elevation} compare the GTSAM variants with existing
legged-IMU odometry baselines on non-elevation-change and elevation-change sequences,
respectively. The compared baselines are
\textbf{Pronto}~\citep{Camurri20frontiers_Pronto},
\textbf{MUSE}~\citep{Nistico25ral_Muse}, and \textbf{Holistic
Fusion}~\citep{Nubert25_Holistic}. Pronto and MUSE are evaluated in proprioceptive-only
configurations without LiDAR or camera input, while Holistic Fusion is evaluated in a
leg-inertial configuration using IMU preintegration, contact-based leg-kinematic landmark
factors, and a custom Spot leg-velocity unary factor from IMU, foot-contact, and
joint-state measurements.

Overall, the GTSAM variants achieve lower accumulated errors than the compared baselines,
especially in translational APE and APE$_z$. This trend is particularly clear in
sequences where contact-induced drift accumulates over long loops, stair traversal, or
slope segments.

On the non-elevation-change sequences (2D), where the motion mostly follows a planar path, the
GTSAM variants consistently reduce translational and vertical drift compared with the
legged-IMU baselines. For example, on CorriLoop and Atrium, the best GTSAM variants
achieve substantially lower translational APE than baselines. Similar trends are observed
in BridgeLoop and Tunnel, where the fixed-lag variants reduce both rotational APE and
APE$_z$. These results indicate that even without exteroceptive sensing, the GTSAM
variants can maintain stable long-term proprioceptive odometry in flat environments.

The benefit becomes more pronounced on elevation-change sequences (3D), where the motion
includes vertical changes. Stair and slope traversal amplify contact uncertainty,
foot slip, and roll--pitch errors, which often appear as vertical drift in purely
legged-IMU odometry. Across Downstair, BiCorridor, Upstair, Quad, Overpass, and
SlopeStair, the GTSAM variants generally achieve the lowest or second-lowest APE$_z$,
showing that the contact-event and fixed-lag formulations are effective for suppressing
vertical drift. Some legged-IMU baselines remain competitive in rotational RPE on a few
sequences, suggesting that their short-horizon orientation estimates can be locally
accurate. However, their accumulated trajectory errors are generally larger, especially
in the vertical direction. These results suggest that the $\SEKThree{K}$-based
invariant filtering representation provides a strong proprioceptive estimation baseline
by yielding a state-independent prediction Jacobian and a structured contact
linearization. The fixed-lag formulations further help suppress accumulated translational and
vertical drift by retaining recent contact episodes and inertial constraints
within a sliding optimization window.

\subsection{Analysis of the Four GTSAM Variants}

\begin{figure}[htbp]
  \centering
  \includegraphics[width=0.75\textwidth]{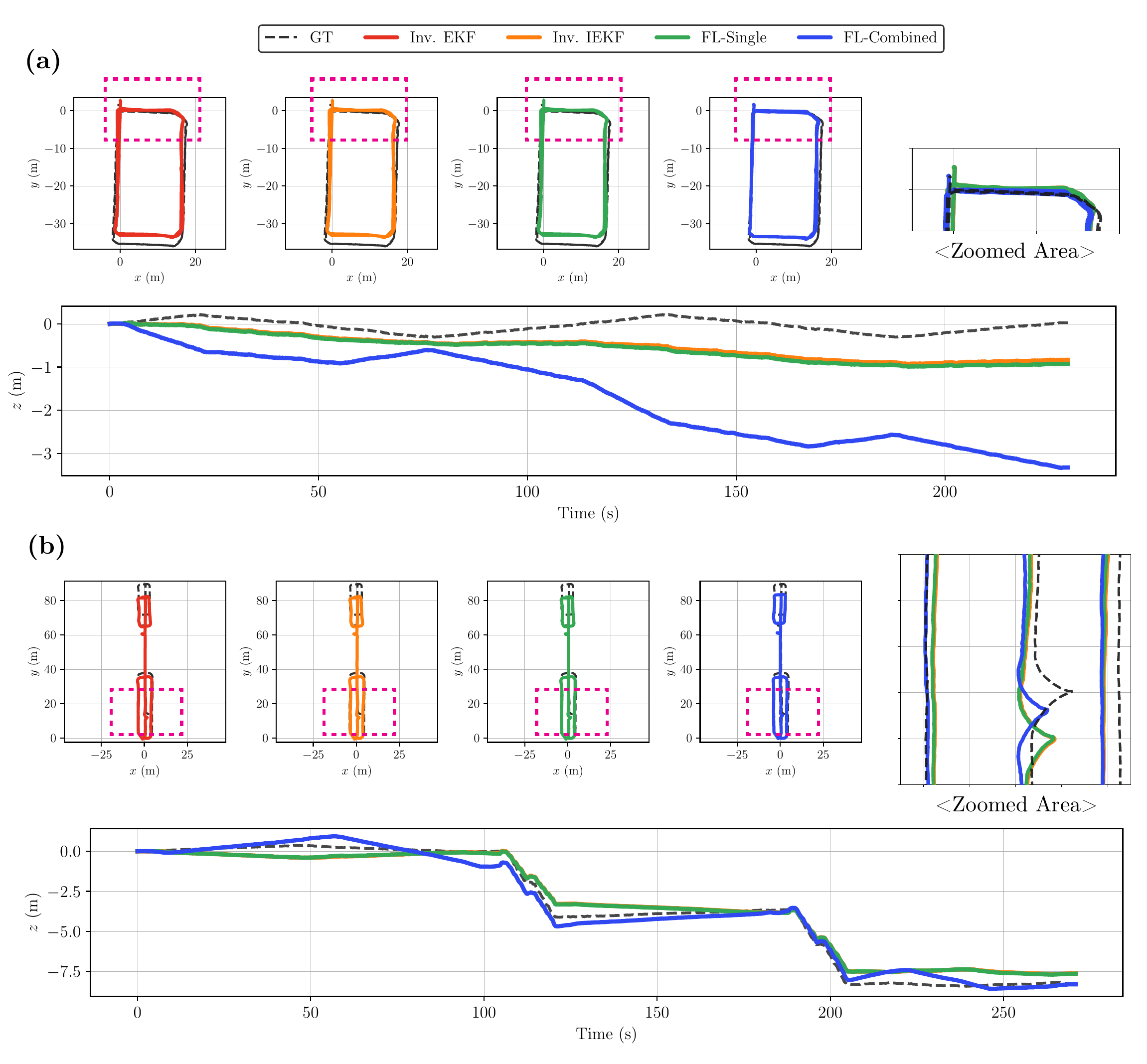}
  \caption{Representative trajectory comparisons of the four GTSAM variants on (a) the
    CorriLoop sequence and (b) the Downstair sequence. The magenta dashed boxes indicate
    regions shown as zoomed-in views within each subplot. While FL-Combined shows the
    closest alignment to the ground truth in the xy plane, the corresponding z-time plots
  reveal that the best xy alignment does not necessarily imply the most accurate vertical estimate.}
  \label{fig:gtsam_variant_trajectories}
\end{figure}

Here we isolate the four GTSAM variants to show the effect of moving from filtering to
graph updates and fixed-lag smoothing. Figure~\ref{fig:gtsam_variant_trajectories}
provides a qualitative comparison on representative non-elevation-change (2D) and
elevation-change (3D) sequences. The CorriLoop result illustrates loop consistency in the
xy plane, while the Downstair result highlights vertical tracking during sustained
elevation change. The plots show that the
variant with the closest xy alignment is not always the one with the lowest vertical
drift, motivating the separate analysis of APE$_t$, APE$_r$, RPE$_t$, RPE$_r$, and
APE$_z$ in the following tables.

\providecommand{\best}[1]{\cellcolor{green!25}\textbf{#1}}
\providecommand{\second}[1]{\underline{#1}}

\begin{table}[htbp]
\centering
\scriptsize
\setlength{\tabcolsep}{4.0pt}
\renewcommand{\arraystretch}{1.08}
\caption{Comparison among GTSAM variants on non-elevation-change (2D) sequences.}
\label{tab:gtsam_invariant_variants_non_elevation}
{%
\begin{tabular}{llccccc}
\toprule
Sequence & Method & APE$_t$ (m) & APE$_r$ (deg) & RPE$_t$ (m) & RPE$_r$ (deg) & APE$_z$ (m) \\
\midrule
\multirow{4}{*}{CorriLoop} & Inv. EKF & \second{2.003} & 1.802 & 0.097 & 0.781 & 0.533 \\
 & Inv. IEKF & \best{1.997} & \second{1.761} & 0.097 & \second{0.781} & \second{0.529} \\
 & FL-Single & 2.010 & 1.815 & \second{0.096} & \best{0.740} & \best{0.525} \\
 & FL-Combined & 2.296 & \best{0.994} & \best{0.096} & 0.808 & 1.776 \\
\midrule
\multirow{4}{*}{Atrium} & Inv. EKF & 1.527 & 1.935 & 0.106 & \second{0.556} & 0.480 \\
 & Inv. IEKF & 1.528 & 1.926 & 0.106 & \best{0.555} & 0.484 \\
 & FL-Single & \second{1.515} & \second{1.790} & \second{0.105} & 0.560 & \second{0.407} \\
 & FL-Combined & \best{0.993} & \best{1.092} & \best{0.090} & 0.565 & \best{0.328} \\
\midrule
\multirow{4}{*}{BridgeLoop} & Inv. EKF & 1.989 & 4.289 & 0.134 & 0.806 & \best{0.793} \\
 & Inv. IEKF & \second{1.988} & \second{4.123} & \second{0.133} & \second{0.803} & 0.809 \\
 & FL-Single & 2.014 & 4.367 & 0.135 & 0.830 & \second{0.802} \\
 & FL-Combined & \best{1.695} & \best{2.218} & \best{0.110} & \best{0.782} & 1.163 \\
\midrule
\multirow{4}{*}{Tunnel} & Inv. EKF & 5.443 & 3.464 & \second{0.086} & 0.441 & \second{1.991} \\
 & Inv. IEKF & 5.483 & 3.460 & 0.087 & \second{0.436} & 2.018 \\
 & FL-Single & \second{5.412} & \second{3.223} & 0.088 & 0.439 & 2.052 \\
 & FL-Combined & \best{4.255} & \best{1.423} & \best{0.079} & \best{0.425} & \best{0.351} \\
\bottomrule
\end{tabular}%
}
\vspace{1mm}
\end{table}

\providecommand{\best}[1]{\cellcolor{green!25}\textbf{#1}}
\providecommand{\second}[1]{\underline{#1}}

\begin{table}[htbp]
\centering
\scriptsize
\setlength{\tabcolsep}{5.0pt}
\renewcommand{\arraystretch}{1.08}
\caption{Comparison among GTSAM variants on elevation-change (3D) sequences.}
\label{tab:gtsam_invariant_variants_elevation}
{%
\begin{tabular}{llccccc}
\toprule
Sequence & Method & APE$_t$ (m) & APE$_r$ (deg) & RPE$_t$ (m) & RPE$_r$ (deg) & APE$_z$ (m) \\
\midrule
\multirow{4}{*}{Downstair} & Inv. EKF & 4.575 & \second{1.585} & 0.133 & 0.713 & \second{1.124} \\
 & Inv. IEKF & 4.591 & 1.592 & 0.133 & 0.711 & 1.130 \\
 & FL-Single & \second{4.544} & \best{1.536} & \second{0.132} & \second{0.679} & \best{1.025} \\
 & FL-Combined & \best{3.842} & 1.829 & \best{0.116} & \best{0.614} & 1.317 \\
\midrule
\multirow{4}{*}{BiCorridor} & Inv. EKF & \second{2.046} & 1.607 & \second{0.099} & 0.680 & \second{0.261} \\
 & Inv. IEKF & \best{2.034} & \best{1.563} & 0.100 & \second{0.678} & 0.263 \\
 & FL-Single & 2.082 & \second{1.605} & 0.100 & 0.684 & \best{0.253} \\
 & FL-Combined & 2.602 & 2.307 & \best{0.099} & \best{0.678} & 2.058 \\
\midrule
\multirow{4}{*}{Upstair} & Inv. EKF & \second{1.460} & 3.219 & 0.117 & 0.750 & \best{0.343} \\
 & Inv. IEKF & \best{1.456} & \second{3.106} & 0.117 & 0.748 & \second{0.350} \\
 & FL-Single & 1.496 & 3.126 & \second{0.116} & \second{0.728} & 0.357 \\
 & FL-Combined & 2.118 & \best{1.496} & \best{0.108} & \best{0.681} & 1.898 \\
\midrule
\multirow{4}{*}{Quad} & Inv. EKF & 6.744 & 2.665 & 0.087 & 0.526 & 2.840 \\
 & Inv. IEKF & \second{6.738} & \second{2.660} & \second{0.087} & 0.526 & 2.846 \\
 & FL-Single & \best{6.670} & \best{2.385} & 0.088 & \best{0.512} & \second{2.581} \\
 & FL-Combined & 8.132 & 3.944 & \best{0.077} & \second{0.522} & \best{2.207} \\
\midrule
\multirow{4}{*}{Overpass} & Inv. EKF & \second{7.627} & 4.331 & \second{0.207} & 0.819 & \second{7.066} \\
 & Inv. IEKF & 7.664 & \second{4.311} & 0.207 & 0.815 & 7.105 \\
 & FL-Single & 7.755 & \best{4.124} & 0.207 & \second{0.772} & 7.150 \\
 & FL-Combined & \best{2.604} & 4.836 & \best{0.110} & \best{0.771} & \best{0.676} \\
\midrule
\multirow{4}{*}{SlopeStair} & Inv. EKF & 3.292 & 1.177 & 0.097 & \second{0.694} & \second{0.590} \\
 & Inv. IEKF & \second{3.282} & \second{1.161} & \second{0.097} & \best{0.693} & 0.590 \\
 & FL-Single & \best{3.248} & 1.164 & 0.098 & 0.721 & \best{0.519} \\
 & FL-Combined & 3.523 & \best{1.127} & \best{0.093} & 0.748 & 2.008 \\
\bottomrule
\end{tabular}%
}
\vspace{1mm}
\end{table}

As presented in Table~\ref{tab:gtsam_invariant_variants_non_elevation} and
Table~\ref{tab:gtsam_invariant_variants_elevation}, across both groups of sequences, Inv.
EKF and Inv. IEKF show very similar performance. This is expected because Inv. IEKF
preserves the same invariant state, prediction model, contact model, and body-frame
foot-error chart as Inv. EKF, while replacing only the measurement correction with a
local nonlinear graph over contact factors at the same time step. The optional height
factors described earlier are disabled in these runs. As a result, the local graph update
gives small improvements in some rotational metrics, but does not
fundamentally change the estimator behavior.

On non-elevation-change sequences (2D), the fixed-lag variants often improve
trajectory-level accuracy. FL-Combined achieves the best APE$_t$ and APE$_r$ on Atrium,
BridgeLoop, and Tunnel, and also gives the best RPE$_t$ on most sequences. This suggests
that allowing the bias to evolve through the lag window can improve accumulated pose
accuracy when the contact and inertial constraints are sufficiently informative. However,
this additional bias flexibility is not uniformly beneficial for APE$_z$: on CorriLoop
and BridgeLoop, FL-Single or the filtering variants provide lower APE$_z$ than
FL-Combined. Thus, the evolving-bias model can improve many accumulated and relative
errors, but its additional degrees of freedom may be less favorable for vertical accuracy
in some sequences.

On elevation-change sequences (3D), the trade-off becomes more visible. FL-Combined is
often strong in RPE$_t$ and RPE$_r$, and it performs particularly well on Overpass, where
it substantially reduces both APE$_t$ and APE$_z$. It also achieves the lowest APE$_t$,
RPE$_t$, and RPE$_r$ on Downstair, although FL-Single gives a lower APE$_z$ on that
sequence. In contrast, on BiCorridor, Upstair, and SlopeStair, the filtering variants or
FL-Single produce better APE$_z$ than FL-Combined. This suggests that, during stair or
slope traversal, the additional bias degrees of freedom may allow the optimizer to
partially absorb contact-induced vertical impacts into the estimated IMU bias, even
though these effects are transient rather than slowly varying bias.
FL-Single is therefore a stable middle ground in several cases: it improves over the
filtering variants in some accumulated metrics while avoiding the larger APE$_z$
sometimes observed with FL-Combined.

\section{Summary}

\vspace{-0.8cm}
\begin{table}[htbp]
\centering
\small
\caption{Summary of the four estimator variants.}
\label{tab:estimator-summary}
{%
\begin{tabularx}{\textwidth}{>{\raggedright\arraybackslash}p{1.05in}
    >{\raggedright\arraybackslash}p{1.75in}
    >{\raggedright\arraybackslash}p{1.15in}
    >{\raggedright\arraybackslash}X}
\toprule
Estimator & State & Update & What changes \\
\midrule
Invariant filter &
Current $\SEKThree{k+2}$ state with body-frame foothold-error blocks &
Sequential \acs{EKF} &
Hartley-style contact-aided \acs{InEKF}; constant contact foot block
$H_f=I$. \\

Graph-update filter &
Same current $\SEKThree{k+2}$ state &
Local nonlinear measurement graph &
Same prediction, but simultaneous contacts are solved jointly. \\

Fixed-lag smoother &
Contact-event base states, persistent bias, navigation-frame foothold
episodes &
Nonlinear fixed-lag graph &
Contact episodes become explicit landmarks and old episodes are
marginalized by the lag window. \\

Combined fixed-lag smoother &
Contact-event poses, velocities, biases, and navigation-frame foothold
episodes &
Nonlinear fixed-lag graph &
Bias evolves through the window via combined preintegration. \\
\bottomrule
\end{tabularx}%
}
\end{table}

We summaize the performance of the four estimator variants in table~\ref{tab:estimator-summary} .
We moved from filtering to graph updates to smoothing
with an explicit bias trajectory: from a single current-state filter, to a local graph
update, to a sliding-window graph, and finally to a sliding-window graph with a bias
trajectory rather than a single persistent bias. The contact side of the graph changes
from current footholds only, to multiple foothold episodes. On the inertial side, the
state expands to a lag rather than the current state, and slowly drifting sensor bias.

Our experiments on the GaRLILEO dataset show that these simple GTSAM variants
provide quite competitive proprioceptive legged odometry without radar, LiDAR, or
camera input. The invariant filtering variants provide stable baselines, while
the fixed-lag variants often improve accumulated trajectory accuracy.

The results also highlight a practical trade-off: additional estimator
expressiveness can improve horizontal or relative accuracy, but it is not
uniformly beneficial for vertical drift across all terrains. In particular,
FL-Combined often improves accumulated or relative errors, whereas FL-Single can
be more stable in APE$_z$ on some elevation-changing sequences. More careful IMU bias
modeling could mitigate this.

\section*{Acknowledgements}

Generative AI tools were used to assist with code generation, figure generation,
and editing of this paper. The authors reviewed and edited all such outputs and
take full responsibility for the content, claims, and results. The factor graph
figures were generated using
\href{https://github.com/dellaert/fgz}{\texttt{dellaert/fgz}}. We also thank all
contributors to GTSAM, past and present.


\appendix

\section{\texorpdfstring{$\SOThree$ Integration Matrices}{SO(3) Integration Matrices}}
\label{app:so3-matrices}

The single-sample filter increment uses the same kernels as
\texttt{so3::DexpFunctor}. Let $\gyro$ be the bias-corrected body angular
velocity used in $U$, and let
\begin{equation}
  \theta = \gyro \Delta t,
  \qquad
  \Omega = \what{\theta},
  \qquad
  \alpha = \|\theta\|.
\end{equation}
The rotation increment is Rodrigues' formula
\begin{equation}
  R(\theta) = I_3 + A\Omega + B\Omega^2,
  \qquad
  A = \frac{\sin\alpha}{\alpha},
  \qquad
  B = \frac{1-\cos\alpha}{\alpha^2}.
\end{equation}
The matrix denoted $J_l$ in the text is
\texttt{DexpFunctor(theta).Jacobian().left()}:
\begin{equation}
  J_l(\theta) = I_3 + B\Omega + C\Omega^2,
  \qquad
  C = \frac{1-A}{\alpha^2}
  = \frac{\alpha-\sin\alpha}{\alpha^3}.
\end{equation}
The matrix denoted $\Gamma_l$ in the text is
\texttt{DexpFunctor(theta).Gamma().left()}:
\begin{equation}
  \Gamma_l(\theta) = \frac{1}{2}I_3 + C\Omega + E\Omega^2,
  \qquad
  E = \frac{1-2B}{2\alpha^2}
  = \frac{\alpha^2 - 2 + 2\cos\alpha}{2\alpha^4}.
\end{equation}
Near zero, we use the corresponding stable series, for
example $C \approx 1/6-\alpha^2/120$ and
$E \approx 1/24-\alpha^2/720$.

\section{Foothold Error Transport}
\label{app:foothold-transport}

We model the footholds as stationary during prediction: the physical foot point
and its nominal estimate both have fixed navigation-frame coordinates,
\begin{equation}
  (f^N_i)^+ = f^N_i,
  \qquad
  (\hat f^N_i)^+ = \hat f^N_i.
\end{equation}
The foot-error Jacobian is nevertheless not the identity. The reason is the
semidirect product structure together with the left-invariant error convention,
i.e., the state perturbation is applied on the right,
\begin{equation}
  X = \hat X\Exp(\xi).
\end{equation}
Thus the foothold tangent block is not an additive navigation-frame error. To
first order in $\xi$, the semidirect product gives
\begin{equation}
  f^N_i = \hat f^N_i + \hat \nRb\delta f^B_i,
\end{equation}
where $\delta f^B_i$ is the right-hand vector block of $\xi$ and is expressed
in the current body frame. After propagation, the same physical point must be
represented in the new right perturbation chart:
\begin{equation}
  (f^N_i)^+ = (\hat f^N_i)^+ + (\hat \nRb)^+(\delta f^B_i)^+,
\end{equation}
up to second-order terms. Equating the two representations and using the
stationarity of $f^N_i$ and $\hat f^N_i$ gives
\begin{equation}
  \boxed{
    (\delta f^B_i)^+ =
    \left((\hat \nRb)^+\right)\T \hat \nRb \, \delta f^B_i.
  }
\end{equation}
Writing $(\hat \nRb)^+ = \hat \nRb\Delta R$, the foothold block in the
prediction Jacobian is therefore
\begin{equation}
  \boxed{
    A_{f_i f_i} =
    \frac{\partial(\delta f^B_i)^+}{\partial \delta f^B_i}
    = \Delta R\T.
  }
\end{equation}

\section{Contact and Front-end Details}
\label{app:contact-front-end}

The invariant contact factor uses the same body-frame contact residual as the
main text. It implements
\begin{equation}
  z^B = (\nRb)\T(f^N-\nPb) + n
\end{equation}
with foot Jacobian block $I$, matching the body-frame foot-error chart. The
smoother contact factors use the same residual, but connect a base variable to
an explicit navigation-frame \texttt{Point3} foothold episode variable. Height
factors act directly on explicit foothold points, and through
$e_3^\top \nRb$ on invariant filter foot blocks.

The contact front end converts measured foot points into the estimator's
body/\acs{IMU} frame before touchdown initialization. Conceptually, the measured
point is first expressed in the body/\acs{IMU} frame used by the estimator, and
touchdown then initializes $\hat f^N = \hat \nPb + \hat \nRb z^B$. The
implementation also supports robust graph contact noise, configured subtractive
\acs{IMU} bias for the filters, and a one-time full-contact initializer before
normal prediction and contact processing begin.

\section{Implementation Map}
\label{app:implementation-map}

This appendix maps the mathematical objects above to the current GTSAM implementation.

\begin{itemize}
  \item $\nRb,\nPb,\nVb$: implemented by \texttt{NavState}. Its pose maps
    body-frame points into the navigation frame.
  \item $\SEKThree{k+2}$ filter state: implemented by
    \texttt{ExtendedPose3(2+k)}. This is the extended-state implementation of
    the semidirect product group. The first two vector blocks are $\nPb$ and
    $\nVb$; the remaining blocks store nominal $f^N_i$. Their tangent
    coordinates are body-frame foot errors.
  \item Sequential invariant filter: implemented by \texttt{LeggedInvariantEKF}
    using \texttt{LeftLinearEKF} with the $X^+=W\phi(X)U$ prediction model and
    sequential contact/height updates.
  \item Local graph update: implemented by \texttt{LeggedInvariantIEKF}. It uses
    the same \texttt{ExtendedPose3(2+k)} state, but replaces the measurement phase by a small
    graph with one prior and the active contact/height factors.
  \item Persistent-bias smoother: implemented by
    \texttt{LeggedFixedLag}\\\texttt{Smoother}. It uses a batch
    fixed-lag smoother, event-based \texttt{NavState} keys, contact-episode
    \texttt{Point3} keys, and one shared bias key.
  \item Evolving-bias smoother: implemented by
    \texttt{LeggedCombined}\\\texttt{FixedLagSmoother}. It uses pose,
    velocity, and bias keys at each event, connected by combined
    preintegration.
\end{itemize}

\newpage

\bibliographystyle{plainnat}
\bibliography{references}

\end{document}